\documentclass[runningheads]{llncs}
\usepackage{graphicx}
\usepackage{caption}
\usepackage{subcaption}
\usepackage{cite}
\usepackage{todonotes}
\usepackage{xcolor}
\usepackage{amsmath}
\usepackage{array,booktabs}
\usepackage{soul}
\usepackage[shortcuts]{extdash}
\newcolumntype{M}[1]{>{\centering\arraybackslash}m{#1}}
\usepackage{hyperref}

\begin{document}
\title{Investigating Premature Convergence in Co-optimization of Morphology and Control in Evolved Virtual Soft Robots
% \thanks{This material is based upon work supported by the National Science Foundation under Grant No. 2008413. Computations were performed on the Vermont Advanced Computing Core supported in part by NSF Award No. OAC-1827314.}
}
\titlerunning{Investigating Premature Convergence}
\author{Alican Mertan\inst{1}\orcidID{0000-0002-5947-8397} \and
Nick Cheney\inst{1}\orcidID{0000-0002-7140-2213} }
\authorrunning{A. Mertan and N. Cheney}
\institute{University of Vermont, Burlington VT 05401, USA\\
\email{\{alican.mertan,ncheney\}@uvm.edu}}

% \author{Anonymous}
% %
% \authorrunning{Anonymous}
% \institute{Anonymous}

%
\maketitle              % typeset the header of the contribution
\begin{abstract}
Evolving virtual creatures is a field with a rich history and recently it has been getting more attention, especially in the soft robotics domain. The compliance of soft materials endows soft robots with complex behavior, but it also makes their design process unintuitive and in need of automated design. Despite the great interest, evolved virtual soft robots lack the complexity, and co-optimization of morphology and control remains a challenging problem. Prior work identifies and investigates a major issue with the co-optimization process -- fragile co-adaptation of brain and body resulting in premature convergence of morphology. In this work, we expand the investigation of this phenomenon by comparing learnable controllers with proprioceptive observations and fixed controllers without any observations, whereas in the latter case, we only have the optimization of the morphology. Our experiments in two morphology spaces and two environments that vary in complexity show, concrete examples of the existence of high-performing regions in the morphology space that are not able to be discovered during the co-optimization of the morphology and control, yet exist and are easily findable when optimizing morphologies alone.  Thus this work clearly demonstrates and characterizes the challenges of optimizing morphology during co-optimization. Based on these results, we propose a new body-centric framework to think about the co-optimization problem which helps us understand the issue from a search perspective. We hope the insights we share with this work attract more attention to the problem and help us to enable efficient brain-body co-optimization.

\keywords{Evolutionary robotics  \and Soft robotics \and Brain-body co-optimization.}
\end{abstract}
\section{Introduction}

Evolving virtual creatures is a field with a long history, starting with Karl Sims' seminal work "Evolving Virtual Creatures"~\cite{sims_evolving_1994} almost 30 years ago. Karl Sims' approach of co-optimizing brain and body then widely adapted and researched, especially in soft robotics~\cite{bhatia2021evolution,joachimczak_artificial_2016,cheney_difficulty_2016,cheney_scalable_2018,medvet_biodiversity_2021,tanaka_co-evolving_2022,mertan_modular_2023,cheney_evolving_2015,cheney_unshackling_2014,creative_machines_lab_cornell_university_ithaca_ny_evolved_2014}. 

The field of soft robotics with volumetric actuation started with~\cite{trimmer_new_2008, hiller_evolving_2010, hiller_automatic_2012} and accelerated with the availability of simulators~\cite{hiller_dynamic_2014, medvet_2d-vsr-sim_2020, bhatia2021evolution,liu2020voxcraft}. The compliance and flexibility of the material make soft robots capable of exhibiting complex and unintuitive behaviors, resulting in different abilities of soft robots such as walking~\cite{creative_machines_lab_cornell_university_ithaca_ny_evolved_2014,cheney_unshackling_2014,joachimczak_artificial_2016,bhatia2021evolution,mertan_modular_2023,cheney_scalable_2018,kriegman_interoceptive_2018,kriegman_how_2018,talamini_evolutionary_2019,kriegman_scalable_2020,medvet_biodiversity_2021,kriegman_scale_2021,medvet_impact_2022,pigozzi_evolving_2022}, swimming~\cite{joachimczak_artificial_2016,corucci_evolving_2016}, squeezing through obstacles~\cite{cheney_evolving_2015}, damage recovery and regeneration~\cite{kriegman_automated_2019,horibe_regenerating_2021,horibe_severe_2022}, shape change~\cite{shah_soft_2021} and self-replication~\cite{kriegman_kinematic_2021} being tested. The highly non-linear and complex nature of soft material dynamics that don't exist in their rigid counterparts~\cite{shepherd2011multigait, rus2015design, kim2013soft}, provides an increased potential for morphological computation~\cite{pfeifer2006body, pfeifer_morphological_2009} that can be unlocked by a tightly integrated body design and control strategy.
 
The ability to output complex behavior, however, makes the design process of soft robots counter-intuitive, highly encouraging automated design over manual design. Despite the great interest in the automated brain-body co-optimization~\cite{hiller_evolving_2010,hiller_automatic_2012,creative_machines_lab_cornell_university_ithaca_ny_evolved_2014,cheney_unshackling_2014,cheney_evolving_2015,cheney_difficulty_2016,corucci_evolving_2016,joachimczak_artificial_2016,cheney_scalable_2018,kriegman_scalable_2020,bhatia2021evolution,talamini_criticality-driven_2021,medvet_biodiversity_2021,kriegman_scale_2021,marzougui_comparative_2022,tanaka_co-evolving_2022,mertan_modular_2023}, evolved soft robots struggles to surpass the complexity of the Sims' initial creatures~\cite{sims_evolving_1994}. Prior work identifies and investigates an important phenomenon that hinders the co-optimization process -- premature convergence of morphology~\cite{joachimczak_artificial_2016,cheney_difficulty_2016}. 

% is is not clear that the following paragraph is the explanation of premature convergence of morphology??

As the optimization takes place, two parts of the solution, the brain and the body, become more and more specialized for each other, making the overall performance of the solution very sensitive to changes in either component~\cite{cheney_difficulty_2016}. In the case of co-optimizing the brain and body in soft robots, this is especially prominent for the morphology of the solution where even small changes in the morphology can drastically reduce the performance of the solution~\cite{mertan_modular_2023}. Commonly used algorithms for brain-body co-optimization don't like solutions with poor performances and focus the search over other parts of the morphology space, making them unable to discover high-performing regions in the search space. To overcome the poor search over morphologies, prior work proposes the use of better search algorithms that reduce the selection pressure of individuals with new body plans to promote more search over morphologies~\cite{lehman_evolving_2011,cheney_scalable_2018}, or proposes ways to alleviate the performance decline during the morphological search by using different genetic representations for coordinated changes~\cite{tanaka_co-evolving_2022,veenstra_effects_2022} or by using controllers that are robust to changes in the morphology~\cite{mertan_modular_2023}.

Rather than proposing a solution, we provide an investigation of this phenomenon following the previous investigation of Cheney et al.~\cite{cheney_difficulty_2016}. Specifically, we investigate the use of neural network controllers with proprioceptive observations where we optimize both morphology and control, and heuristically devised fixed controllers where we only optimize the morphology. This allows us to shed light on the effects of control optimization and specialization on the co-optimization process and understand the premature convergence issue. The main contributions of this work are to:
\begin{itemize}
    \item expand previous investigations towards a more generalized understanding of the brain-body fragile co-optimization phenomenon by focusing on important and missing aspects such as the use of more varied and complex controllers, sensory information, morphology spaces, and environments. (Sec.~\ref{sect:co-op})
    \item provide arguably the most concrete example that shows the existence of high-performing designs that aren't discovered by the co-optimization process, due to poor search over the morphology space. (Sec.~\ref{sect:analysis})
    \item develop a new body-centric framework to conceptualize the fitness landscape of the co-optimization problem, allowing us to understand the challenge of co-optimization from a search perspective. (Sec.~\ref{sect:discussion})
    \item organize the existing solutions to premature convergence within the proposed framework, helping us to clarify promising research directions. (Sec.~\ref{sect:discussion})
\end{itemize}
We hope these findings and the proposed framework will be useful for further research into enabling better brain-body co-optimization.

% \section{Background}

% \input{sections/background}

\section{Experiment Design}

\subsection{Simulation}
We perform our experiments in the Evolution Gym version 1.0.0 (Evogym)~\cite{bhatia2021evolution}. It is an open-source voxel-based soft-body simulator with a suite of benchmark tasks. Voxels are represented as a mass-spring system, and it works in 2D, similar to the simulation engines in~\cite{ferigo_evolving_2021, joachimczak_artificial_2016, medvet_evolution_2020, medvet_2d-vsr-sim_2020, medvet_biodiversity_2021, medvet_impact_2022, pigozzi_evolving_2022, tanaka_co-evolving_2022, mertan_modular_2023}. 

Each voxel of the robot can have a different type that determines its behavior and is represented with different colors visually. There are two types of active materials that can actuate volumetrically, either \textcolor{orange}{horizontally} or \textcolor{cyan}{vertically}, and they are under the direct control of the controller. 
These materials are controlled by specifying the target length $a \in [0.6,1.6]$ (in the form of a multiplier of the resting length) at each time step where materials gradually expand/contract to achieve the target length. We query the controller every 5\textsuperscript{th} timestep (referred to as the effective timestep) and repeat the queried action until the next effective time step to limit high-frequency dynamics that might lead to unstable behavior. Additionally, there are two types of passive materials that are not under direct control, and they differ in their elasticity, one being \textbf{rigid} and the other being \textcolor{gray}{elastic}. Designing a robot is simply choosing the existence and material type of voxels in a grid layout. While Evogym allows for specifying whether neighboring voxels are connected to each other, we omit this feature and assume that neighboring voxels are always connected to each other.

The simulation engine provides various proprioceptive and environmental observations. We only use proprioceptive observations of volume, speed, and material properties of voxels and a global time signal. 

\subsection{Task and Environments}
We use two environments with a locomotion task that are presented in the Evogym benchmark suite, namely Walker-v0 and BridgeWalker-v0~\cite{bhatia2021evolution}. Locomotion is the most used task for voxel-based soft robotics research~\cite{mertan_modular_2023,creative_machines_lab_cornell_university_ithaca_ny_evolved_2014,cheney_unshackling_2014,cheney_difficulty_2016,cheney_scalable_2018,talamini_evolutionary_2019, medvet_evolution_2020,talamini_criticality-driven_2021,medvet_biodiversity_2021,medvet_impact_2022,horibe_regenerating_2021,kriegman_interoceptive_2018,marzougui_comparative_2022,horibe_severe_2022,tanaka_co-evolving_2022,corucci_evolutionary_2017,kriegman_how_2018,kriegman_automated_2019,kriegman_scalable_2019,kriegman_scale_2021,talamini_evolutionary_2019,ferigo_evolving_2021,pigozzi_evolving_2022,cheney_unshackling_2014}. We adopt the modified fitness function in~\cite{mertan_modular_2023}, $R(r,T) = \Delta p_x^r + \mathbf{I}(r) + \sum_{t=0}^T -0.01 + 5$, which rewards robots $r$ for moving in the positive $x$ direction with the term $\Delta p_x^r$, rewards them for finishing the task with the indicator function $\mathbf{I}(p_x^r)$, and encourages them to finish the task faster by applying a negative penalty at each time step with the term $\sum_{t=0}^T -0.01$. The last term, $+5$, is equal in magnitude to the maximum penalty and is used to shift the reward to be positive for ease of analysis. 

Environments often consist of a flat~\cite{mertan_modular_2023,creative_machines_lab_cornell_university_ithaca_ny_evolved_2014,cheney_unshackling_2014,cheney_difficulty_2016,cheney_scalable_2018,talamini_evolutionary_2019, medvet_evolution_2020,talamini_criticality-driven_2021,medvet_biodiversity_2021,medvet_impact_2022,horibe_regenerating_2021,kriegman_interoceptive_2018,marzougui_comparative_2022,horibe_severe_2022,tanaka_co-evolving_2022,corucci_evolutionary_2017,kriegman_how_2018,kriegman_automated_2019,kriegman_scalable_2019,kriegman_scale_2021} or uneven~\cite{talamini_evolutionary_2019,ferigo_evolving_2021,pigozzi_evolving_2022,cheney_unshackling_2014} surfaces, supposedly harder task. We experiment with both locomotion on a flat surface (Walker-v0) and an uneven, dynamic surface (BridgeWalker-v0) to understand the relationship between the environment and the brain-body co-optimization process.

\subsection{Robot Design and Controller Strategies}

\subsubsection{Robot Representation and Design} Following~\cite{bhatia2021evolution,mertan_modular_2023,medvet_biodiversity_2021}, we use a direct representation for robot design where the existence of voxels and their material type encoded in a matrix $\in M_{h \times w}(T)$, where $T \in$ \{0,\textbf{1},\textcolor{gray}{2},\textcolor{orange}{3},\textcolor{cyan}{4}\}. 
While indirect encodings are commonly used in prior work that works in 3D~\cite{creative_machines_lab_cornell_university_ithaca_ny_evolved_2014,cheney_unshackling_2014,cheney_evolving_2015,cheney_difficulty_2016,corucci_evolving_2016,joachimczak_artificial_2016,cheney_scalable_2018,kriegman_interoceptive_2018,kriegman_scalable_2020}, working in 2D makes it possible to use direct encoding which is shown to be comparable to complex indirect encodings~\cite{bhatia2021evolution} and allows us to control for the change in morphology during mutation. Following the common practice of limiting the morphology space~\cite{creative_machines_lab_cornell_university_ithaca_ny_evolved_2014, cheney_difficulty_2016, cheney_scalable_2018, kriegman_how_2018, medvet_biodiversity_2021, marzougui_comparative_2022, tanaka_co-evolving_2022,mertan_modular_2023}, we experiment with two morphology spaces, $(h,w) \in \{(5,5), (7,7)\}$, to provide more instances for investigation of premature convergence, as encouraged by previous investigation~\cite{cheney_difficulty_2016}.

\subsubsection{Controller Design and Model} 
We compare a learnable controller and a fixed, non-sensing controller. The fixed controller allows us to create a scenario where fragile co-adaptation of design and controller cannot occur due to the lack of optimization on the controller part and helps us investigate premature convergence that we observe with learnable controllers. Specifically, we use a modular/decentralized control strategy as a learnable control
\footnote{We repeated every experiment with the most commonly used global/centralized control strategy~\cite{medvet_evolution_2020, huang_one_2020, ferigo_evolving_2021, mertan_modular_2023, talamini_evolutionary_2019, talamini_criticality-driven_2021,medvet_impact_2022} and didn't observe any qualitatively different results. For the sake of space and simplicity, we omit the global controller and only present the results with the modular controller.}
as they are shown to help with the brain-body co-optimization~\cite{mertan_modular_2023}. 

\textit{Learnable Controller} One of the common choices in control design is the use of modular control strategy~\cite{pigozzi_evolving_2022, medvet_evolution_2020, medvet_biodiversity_2021, huang_one_2020,mertan_modular_2023}. They are considered to be more versatile and robust to changes in the robot morphology~\cite{yim_modular_2007}, which helps with the co-optimization process~\cite{mertan_modular_2023}, and are compatible with any robot morphology.

The modular controller consists of a shared neural network model (single hidden layer with 32 neurons) assigned to each active voxel, observing a local patch around it and only determining the behavior of a single voxel it is assigned. The controller observes a $3\times3$ window centered around the active voxel. The observations to the controller consist of the proprioceptive information (volume, speed, and material type) of the voxels in this observation window and $\text{timestep}\bmod 2$ as the time signal to have a fair comparison with the fixed controller. 

% Modular controllers are modeled with a one-hidden layer of 32 neurons, multi-layer perceptron with $2401$ parameters. 
% Note that the controller is compatible with any robot body due to modularity and the parameters do not change with the body or the morphology space, since the modular controller works with local observations.

\textit{Fixed Controller} One of the simplest control strategies used in the literature is the fixed actuation following a global signal~\cite{cheney_unshackling_2014,cheney_evolving_2015,cheney_difficulty_2016,corucci_evolving_2016,cheney_scalable_2018,kriegman_interoceptive_2018,kriegman_how_2018,kriegman_automated_2019,kriegman_scalable_2019,kriegman_scalable_2020,kriegman_scale_2021,talamini_criticality-driven_2021}. While the learnable controller performs a nonlinear mapping from proprioceptive observations to actions, fixed controller lacks any sensory input. Active voxels under the control of the fixed controller alternate between expanding and contracting maximally at every effective timestep. Therefore, the controller has zero learnable parameters and doesn't require any control optimization, hence the name fixed. Having a fixed controller that doesn't require any control optimization allows us to create a scenario where fragile co-adaptation doesn't occur, helping us understand premature convergence with learnable controllers better. 

\subsection{Optimization Algorithm}
To optimize the design and control of the soft robots, we adopt the use of evolutionary algorithms, similar to~\cite{joachimczak_artificial_2016,cheney_difficulty_2016,cheney_scalable_2018,veenstra_effects_2022,veenstra_how_2020,pontes-filho_single_2022,medvet_biodiversity_2021,tanaka_co-evolving_2022,mertan_modular_2023}. Evolutionary algorithms allow us to simultaneously make improvements in both design and control, supposedly reducing the computational cost of two-level optimization approaches where outer loop does design optimization and the inner loop optimizes the control for a given design, as in~\cite{bhatia2021evolution}.

In particular, we use age-fitness Pareto optimization (AFPO)~\cite{schmidt_age-fitness_2010} with truncation selection. Individuals' ages are increased at every generation, and we inject a random individual at each generation with the age of $0$ for diversity. Recombination wasn't considered and the offspring are created through mutation only. Individuals created by mutation inherit the age of their parent.

Following~\cite{bhatia2021evolution,mertan_modular_2023}, we use a mutation operator that creates new designs by changing each voxel type of a robot with a $10\%$ probability from/to an empty voxel. To mutate in the control space, we add noise sampled from $\mathcal{N}(0,0.1)$ to all learnable controller parameters. 

When creating offspring through mutation, we either mutate the body plan or the controller of the solution. Following~\cite{cheney_scalable_2018,mertan_modular_2023}, there is a $50\%$ probability of choosing either component of the solution for the mutation. Note that the fixed controller doesn't have any controller parameters. Yet, for a fair comparison, individuals with fixed controllers go through control mutation with the same probability, effectively creating a copy of the same solution. 

\section{Co-optimization}
\label{sect:co-op}

\begin{figure}[t]
    \centering
    \begin{subfigure}{0.5\textwidth}
            \includegraphics[width=\textwidth]{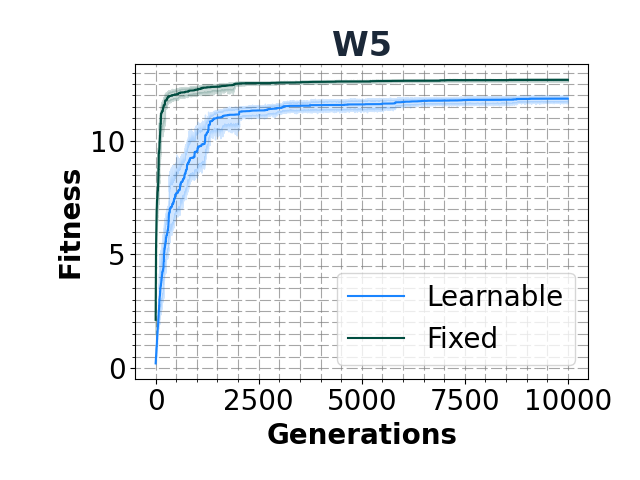}%
    \end{subfigure}%
    \begin{subfigure}{0.5\textwidth}
            \includegraphics[width=\textwidth]{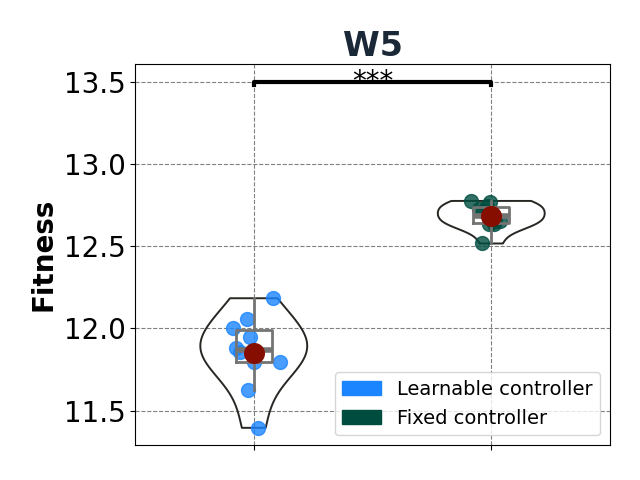}%
    \end{subfigure}\\
    \begin{subfigure}{0.5\textwidth}
        \includegraphics[width=0.33\textwidth]{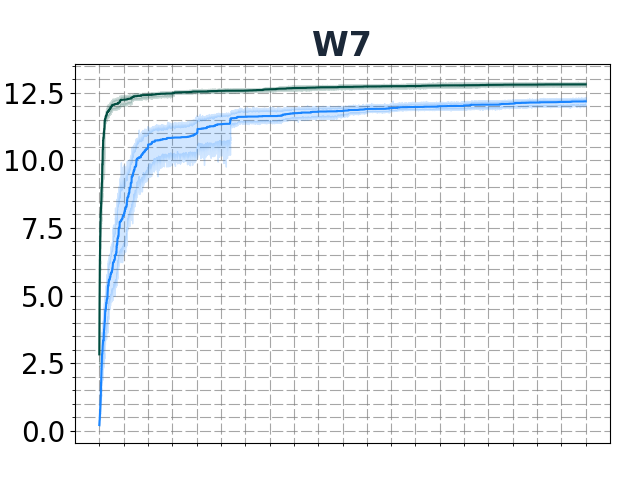}%
        \includegraphics[width=0.33\textwidth]{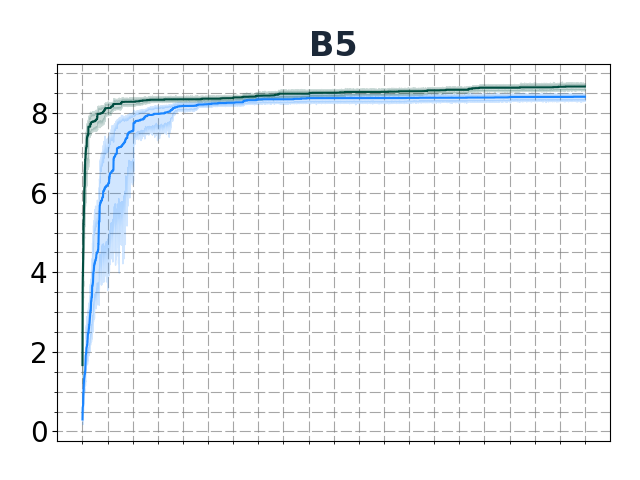}%
        \includegraphics[width=0.33\textwidth]{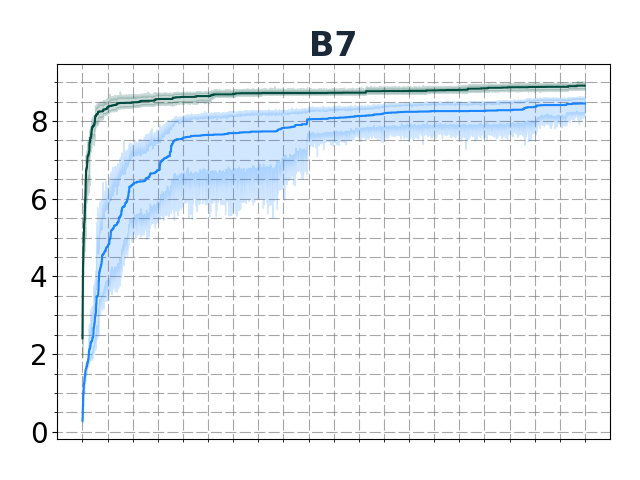}%
    \end{subfigure}%
    \begin{subfigure}{0.5\textwidth}
        \includegraphics[width=0.33\textwidth]{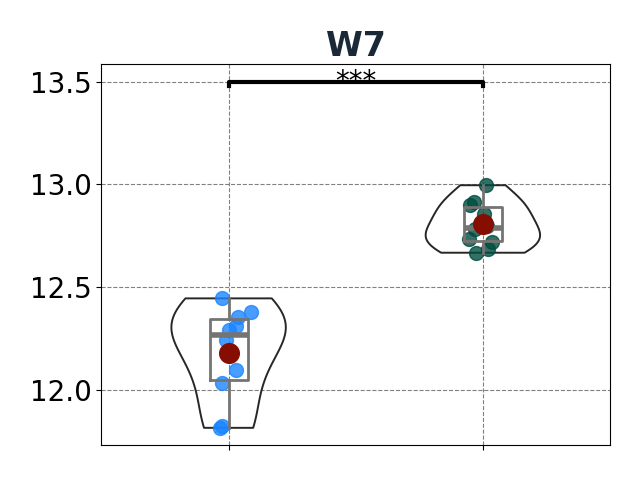}%
        \includegraphics[width=0.33\textwidth]{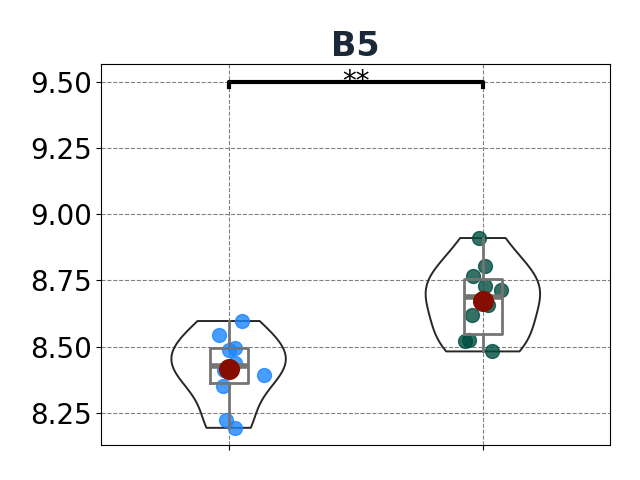}%
        \includegraphics[width=0.33\textwidth]{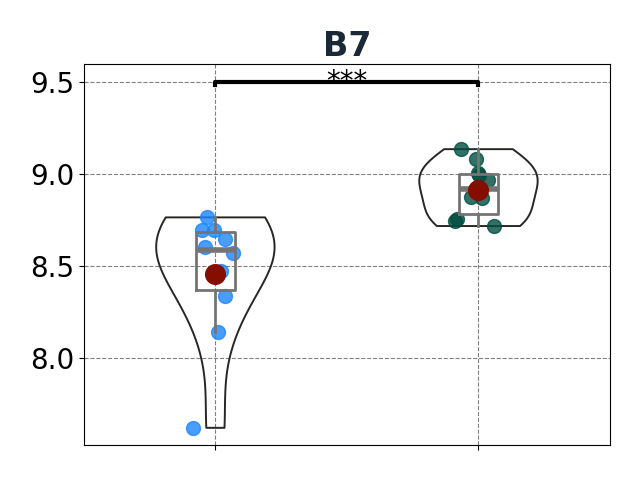}%
    \end{subfigure}%
    \caption{Fitness over time plots (left) and distributions of the best solutions (right) for the co-optimization experiments.
    % (W: Walker-v0, B: BridgeWalker-v0, 5: $(h,w)=(5,5)$, 7: $(h,w)=(7,7)$). 
    The simplest controller, the fixed controller, significantly outperforms the learnable controller in all experimental settings. Moreover, populations with fixed controllers converge faster.
    Left: Solid lines show the best fitness found at each generation, averaged across 10 runs. Shaded regions show the $95\%$ confidence intervals. 
    Right: Each data point is plotted, as well as the mean values which are marked with dark red. Horizontal lines indicate statistically different results. ***: $P<0.001$, **: $P<0.005$, *: $P<0.05$}
    \label{fig:fots}
\end{figure}

The main problem we are interested in is the co-optimization of morphological design and control, and understanding the premature convergence of morphology. To compare the effects of different control strategies, we evolve two populations of solutions: one with learnable and one with fixed controllers.
While with the learnable controller, we have the co-optimization of morphology and control, the fixed controller allows us to test the case where we only optimize the morphology, allowing us to investigate the effect of controllers on the premature convergence issue.
We experiment with evolving these populations under 4 different settings by varying the environment, Walker-v0 and BridgeWalker-v0, and the morphology space, $(h,w) \in \{(5,5), (7,7)\}$. Each setting is named as \{\textbf{E}nvironment name\}\{(\textbf{h},w)\}, e.g. \textbf{W5} for the setting \textbf{W}alker-v0,(\textbf{5},5). For each setting, we repeat the experiment 10 times and use the Wilcoxon Rank Sum test~\cite{wilcoxon1964some} to report $P$-values. Each run consists of a population size of 16 individuals evolved for 10000 generations with the AFPO algorithm~\cite{schmidt_age-fitness_2010}.

Figure~\ref{fig:fots} left, illustrates the fitness over evolutionary time plots for each experiment. Solid lines show the fitness of the best solution found at each generation, averaged across 10 runs. Shaded regions show the $95\%$ bootstrapped confidence interval. Additionally, Figure~\ref{fig:fots} right illustrates the distribution of the performance of the final solutions. The results demonstrate that the populations with fixed controllers find better solutions faster~(all $P < 0.05$ for achieving $85\%$ of final performance) and find significantly better solutions~(all $P < 0.005$). These results are consistent across different environments and morphology spaces. While individuals with fixed controllers are, admittedly, incapable of observing their environment, hence unlikely to scale to more complex problems, \textit{fixed controllers nevertheless provide a strong baseline for performance for simple tasks such as locomotion, where a lot of research is still being done.}

The success of fixed controller might seem interesting, yet it is a well-established phenomenon that it is difficult to optimize body plan and control together~\cite{cheney_difficulty_2016,mertan_modular_2023,cheney_scalable_2018,joachimczak_artificial_2016}.
Having two interdependent parts of a solution creates a challenging optimization problem where the parts become more and more specialized for each other as the optimization takes place, making them unamenable to change without breaking the solution's performance -- fragile co-adaptation. This is especially true for morphology, as it is the interface between the controller and environment and thus changing morphology also scrambles the controller. It makes it harder to search the morphology space, resulting in premature convergence of morphology~\cite{cheney_difficulty_2016,joachimczak_artificial_2016}.
On the other hand, the lack of parameters for the fixed controller turns the co-optimization of the morphology and control problem into only the optimization of the morphology, eliminating the issue of fragile co-adaptation. The results demonstrating superior performance for fixed controllers suggest that \textit{fixed controller doesn't suffer from premature convergence and allows optimization process to find body plans that outsource complexity to the dynamics of the system, an example of morphological computation~\cite{pfeifer_morphological_2009}.}

% Yet it isn't obvious that only optimizing the design with fixed actuation would yield high-performing solutions at all. Indeed, our results might seem at odds with the results of Talamini et al.~\cite{talamini_evolutionary_2019}, where global controllers are shown to be more effective than non-sensing controllers with fixed actuation. However, we would like to point out that Talamini et al.~\cite{talamini_evolutionary_2019} experiment with arbitrarily chosen fixed morphologies while we allow an optimization process to take place for the design. This discrepancy, and the fact that fixed controllers outperformed learnable controllers in our experiments, indicate that \textit{certain morphologies require more complex control strategies/gaits,} e.g.\ the ones used in Talamini et al.~\cite{talamini_evolutionary_2019}, \textit{while others can perform well with fixed actuation}, e.g.\ the morphologies discovered in our experiment. We conjecture that \textit{fixed controller allows optimization process to find designs that outsource complexity to the dynamics of the system, an example of morphological computation~\cite{pfeifer_morphological_2009}.} We provide more evidence for this claim in Section~\ref{sect:analysis}.

Lastly, we note that the fixed controller consistently outperforms the more complex learnable controller in the BridgeWalker-v0 environment in two different morphology spaces. In light of these results, \textit{we conjecture that the locomotion over uneven surface doesn't present fundamentally harder/different optimization problem than locomotion on flat surface for brain-body co-optimization.}

\section{Analysis} \label{sect:analysis}

Our experiments demonstrate the inferior performance of co-optimization with learnable controllers compared to morphology optimization with fixed controllers. In this section, we perform a series of analyses to understand how the differences in controllers affect the co-optimization. Since we expect learnable controllers to be capable of learning to output the simple behavior that the fixed controllers exhibit, the differences in performance should arise from the different controllers' effects on the search over the morphology space. As prior work strongly emphasizes the issue of premature convergence as the major hurdle in the brain-body co-optimization~\cite{cheney_difficulty_2016,joachimczak_artificial_2016,mertan_modular_2023}, we specifically focus on this issue in our analysis.

\subsubsection{Qualitative analysis of body plans}

As we discussed earlier, one of the biggest challenges of brain-body co-optimization is the premature convergence of the morphology~\cite{cheney_difficulty_2016,joachimczak_artificial_2016,mertan_modular_2023}. While optimizing a controller for a given body plan is relatively easy, search over the morphology space often prematurely converges to a local optimum. To compare experimented controllers in terms of their effect on the search over the morphology space, we start by looking at the body plans of the best solutions found at each run, i.e.\ run champions, under each setting.

\begin{figure}[t]
    \centering
    \includegraphics[width=0.5\textwidth]{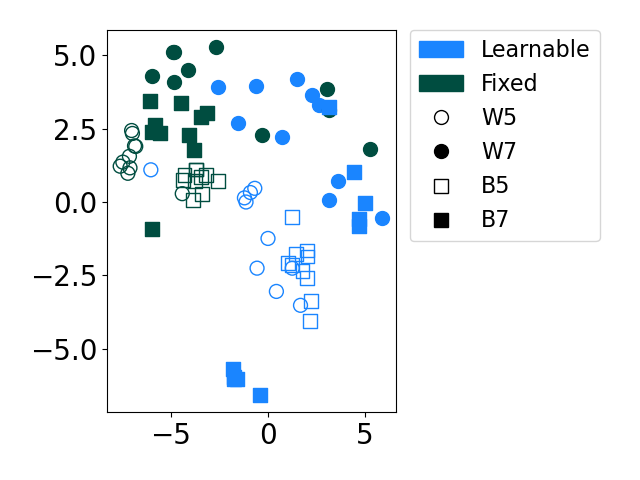}%
    \includegraphics[width=0.5\textwidth]{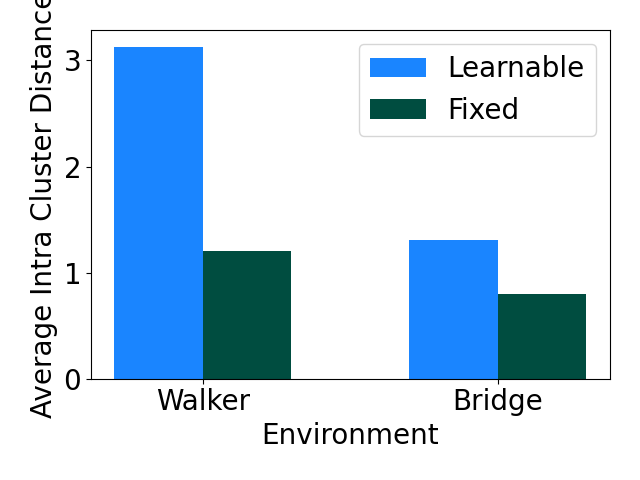}
    \caption{t-SNE plots of the run champions' body plans and average intra-cluster distances in the embedding space. The populations with the fixed controller converge better to a body plan across different runs and morphology spaces.}
    \label{fig:tsne}
\end{figure}

First, we try to measure the run champions' convergence to a particular body plan, which indicates a better search over the morphology space as we expect a successful search to be able to exploit high-performing regions of the space across runs~\cite{cheney_difficulty_2016}. To this end, we perform t-SNE dimensionality reduction~\cite{van2008visualizing} of the optimized morphologies.  We visualize the clusters in Figure~\ref{fig:tsne} and measure the average intra-cluster distance, to quantify the variation in found morphologies for a given controller paradigm. We find that run champions with fixed controllers converge to more similar solutions across different runs and morphology spaces in two different environments.
We also note that the learnable run champions show less diversity in the BridgeWalker-v0 environment, showing that the environment also plays a role in shaping the fitness landscape and affecting which body plans can be easily found during the co-optimization process.

\begin{figure}[!hp]
    \centering
    \begin{subfigure}{0.125\textwidth}
        \centering
        \includegraphics[width=0.7\textwidth]{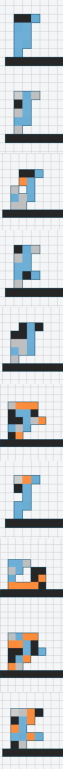}%
        \caption{L-W5}
    \end{subfigure}%
        \begin{subfigure}{0.125\textwidth}
        \centering
        \includegraphics[width=0.7\textwidth]{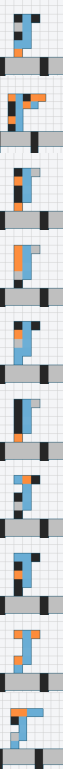}%
        \caption{L-B5}
    \end{subfigure}%
    \begin{subfigure}{0.125\textwidth}
        \centering
        \includegraphics[width=0.7\textwidth]{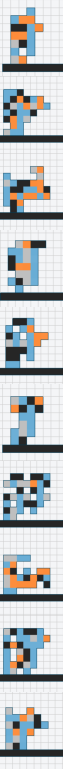}%
        \caption{L-W7}
    \end{subfigure}%
    \begin{subfigure}{0.125\textwidth}
        \centering
        \includegraphics[width=0.7\textwidth]{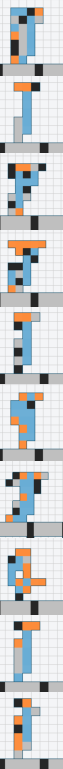}%
        \caption{L-B7}
    \end{subfigure}%
    \vrule
    \begin{subfigure}{0.125\textwidth}
        \centering
        \includegraphics[width=0.7\textwidth]{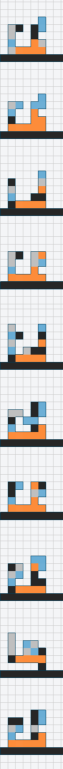}%
        \caption{F-W5}
    \end{subfigure}%
        \begin{subfigure}{0.125\textwidth}
        \centering
        \includegraphics[width=0.7\textwidth]{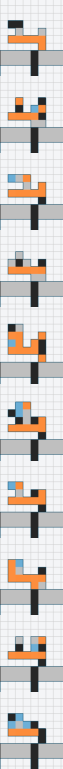}%
        \caption{F-B5}
    \end{subfigure}%
    \begin{subfigure}{0.125\textwidth}
        \centering
        \includegraphics[width=0.7\textwidth]{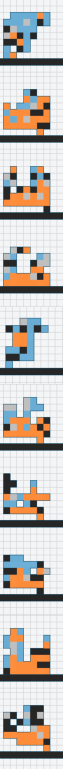}%
        \caption{F-W7}
    \end{subfigure}%
    \begin{subfigure}{0.125\textwidth}
        \centering
        \includegraphics[width=0.7\textwidth]{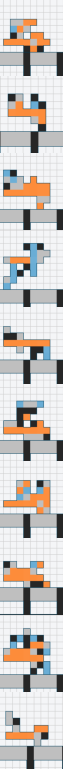}%
        \caption{F-B7}
    \end{subfigure}%
    \caption{Morphologies of the run champions under each experimental setting. 
    The left half of the figure shows the run champions with learnable controllers, and the right half shows the ones with fixed controllers. 
    We see diverse body plans evolve with the learnable controller, especially in (a). On the other hand, we see the same form of the mostly active bottom with two vertical apparatus at the front and back, resembling a head and a tail, evolved with the fixed controller in (e). For the BridgeWalker-v0 environment, individuals with learnable controllers show similar body plan features in (b) and (d), e.g.\ upright posture, r or T shape, individuals with fixed controller usually consists of a thin, horizontal body with active materials and forward apparatus resembling a leg in (f-h).}
    \label{fig:runchamps}
\end{figure}
\begin{figure}[!hp]
    \centering
    \begin{subfigure}{\textwidth}
        \includegraphics[width=\textwidth]{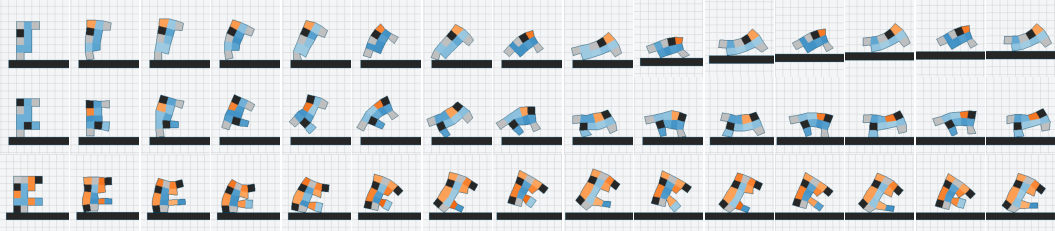}
        
        \vspace*{1em}
        
        \includegraphics[width=\textwidth]{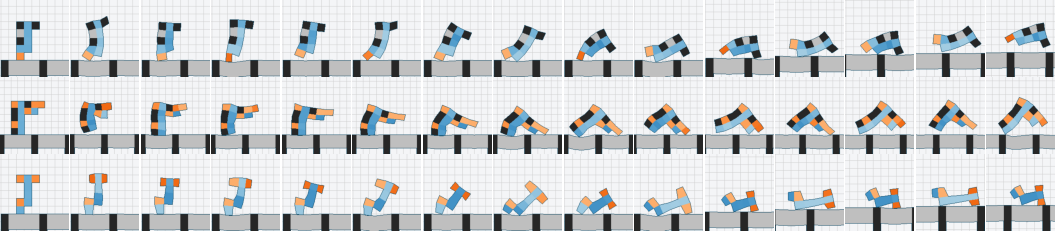}
        \caption{Learnable Controller}
    \end{subfigure}
    \begin{subfigure}{\textwidth}
    
        \vspace*{1em}
    
        \includegraphics[width=\textwidth]{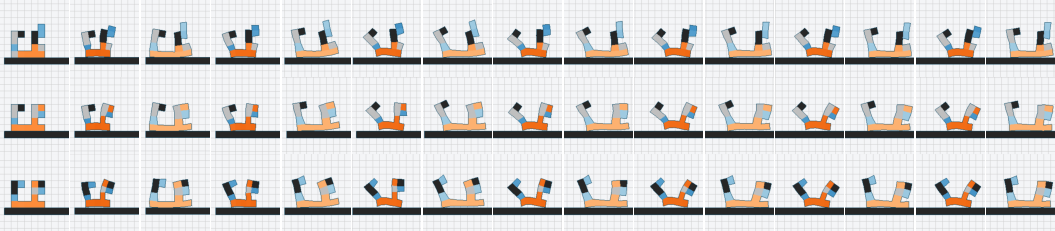}
        
        \vspace*{1em}
        
        \includegraphics[width=\textwidth]{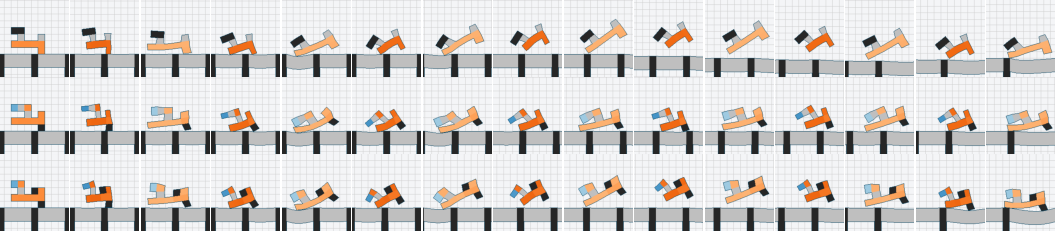}
        \caption{Fixed Controller}
    \end{subfigure}
    \caption{Timelapse images of three run champions' behavior from each of the two tasks of the $(5,5)$ morphology space, co-optimized brain and body (a) and morphology-only optimization with fixed control (b). 
    % The inductive bias of the controllers, with the combination of the given task, environment, and optimization algorithm, shapes the search space in certain ways, allowing populations to find similar designs in certain cases and not in others. 
    Learnable controllers demonstrate more diversity, especially in the Walker-v0 environment. Most of the individuals with learnable controllers (a) find a bipedal (rows 1, 2, 4-6) or monopedal gait (row 3).
    % Bipedal gaits also show variation. Some individuals achieve bipedal locomotion by actuating different muscles out of phase (row 2), while others use their muscles mostly in-phase (rows 1, 4-6).  
     Fixed controllers (b) show one common behavior on flat ground (rows 1-3), consisting of an active bottom with two vertical apparatus at the front and back, resembling a head and a tail. When the muscle at the bottom of the robot contracts, vertical apparatuses help the bottom part form an arch, creating front and back legs to locomote.  In the bridge environment (rows 4-6) often includes a thin, horizontal body with active materials, a forward apparatus resembling a leg, and some upper body (presumably used for balancing). Individuals with this body plan throw themselves forward by actuating their muscles in phase and pulling themselves forward with their forward apparatus, achieving bipedal locomotion.
    }
    \label{fig:timelapse}
\end{figure}

To qualitatively analyze the robot body plans and gaits, we present Figure~\ref{fig:runchamps}, where we show the body plans of the run champions, and Figure~\ref{fig:timelapse}, where we display 15 snapshots during the lifetime of select individuals. 
% Additionally, some individuals start locomotion by falling forward initially (Figure~\ref{fig:timelapse} (a): rows 1, 2, 4-6), a strong local optima that is found early on in the optimization process, some individuals preserve their initial orientation (Figure~\ref{fig:timelapse} (a): row 3).
In our experimental setup, fixed controllers (Fig.~\ref{fig:timelapse}b) consistently find similar body plans with near identical gaits. We do not observe this level of converge across runs that co-optimize morphology with a learnable controller (Fig.~\ref{fig:timelapse}a).   This trend holds across all experimental settings.
% On the other hand, run champions with fixed controllers show more similarity in design and gait. This is especially true for champions found in the same experimental setting. Yet, we also see more similarities across different environments and design spaces as well. 
% The first design (prominent in Figure~\ref{fig:runchamps} (b) and Figure~\ref{fig:timelapse} (b): rows 1-3) consists of an active bottom with two vertical apparatus at the front and back, resembling a head and a tail. When the muscle at the bottom of the robot contracts, vertical apparatuses help the bottom part form an arch, creating front and back legs to locomote. The second design (prominent in Figure~\ref{fig:runchamps} (d), and in Figure~\ref{fig:timelapse} (b): rows 4-6) consists of a thin, horizontal body with active materials, forward apparatus resembling a leg, and some upper body (presumably used for balancing). Individuals with this design throw themselves forward by actuating their muscles in phase and pulling themselves forward with their forward apparatus, achieving bipedal locomotion.

\subsubsection{Quantitative analysis of body plans}

Our qualitative analysis of body plans shows that champions with the learnable controller show more diversity while fixed run champions are grouped closer in the morphology space. The diversity of solutions found with learnable controllers can be interpreted as a positive feature. However, the literature strongly suggests that it is a sign of poor search over the morphology space~\cite{cheney_difficulty_2016}. Although it is possible that the search space contains many distinct near-optimal solutions, the high-performing body plans of fixed run champions imply the existence of better-performing body plans that aren't discovered in runs with the learnable controller. 

Therefore in this section, we try to definitively answer our main question: \ul{have learnable controllers failed to find high-performing body plans that were discovered by fixed controllers?} To answer this question, we take the morphologies of the run champions found in runs with fixed controllers and no co-optimization, then optimize a learnable controller for each body plan.   We suggest that by pairing the same learned closed-loop controller from the co-optimization paradigm to the morphologies found when optimized with the fixed controller, we will be able to fairly compare how well each optimization setup is able to find high-performing morphologies in the search space.  
% demonstrates the superiority of designs found with the fixed controller, even when they are controlled by the learnable controller. 
We take the body plans of the run champions with fixed controllers and start new evolutionary runs from scratch, with 16 individuals having the same fixed body plan as the corresponding run champion, and optimize only the controller 
% with age-fitness Pareto optimization~\cite{schmidt_age-fitness_2010} 
for 5000 generations.   

\begin{figure}[t]
    \centering
    \begin{subfigure}{0.7\textwidth}
        \includegraphics[width=\textwidth]{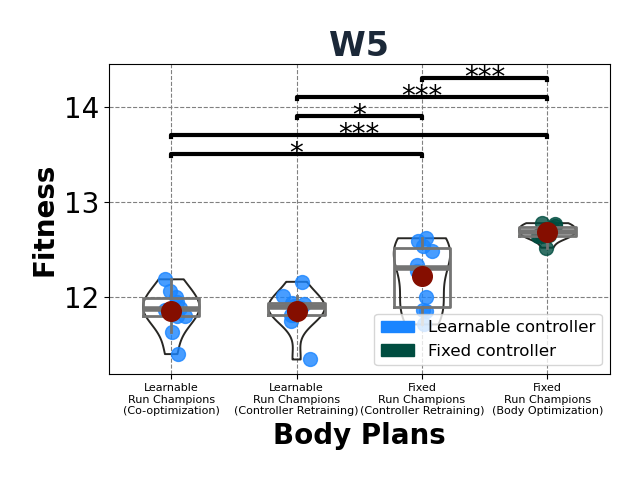}%
    \end{subfigure}%
    \begin{subfigure}{0.3\textwidth}
        \includegraphics[width=0.8\textwidth]{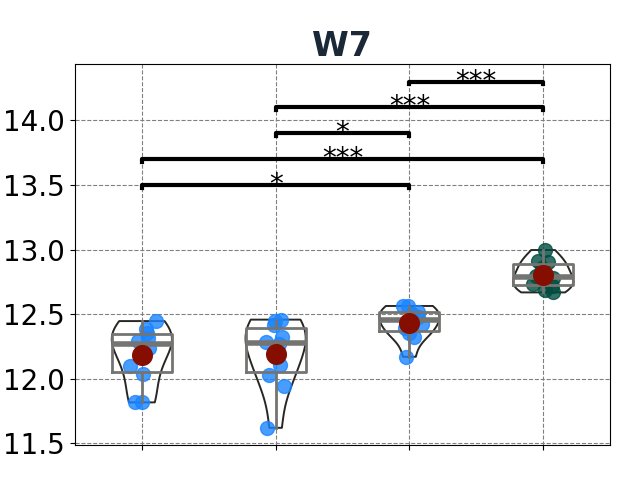}\\
        \includegraphics[width=0.8\textwidth]{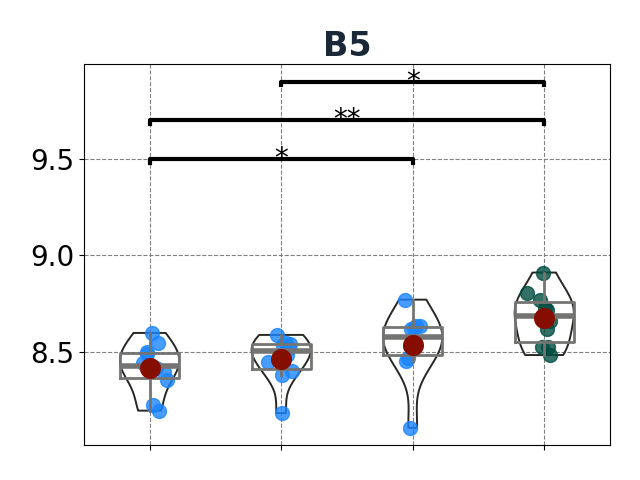}\\
        \includegraphics[width=0.8\textwidth]{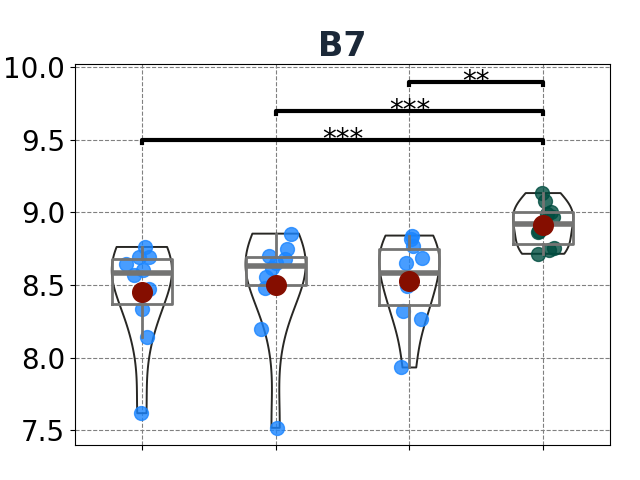}
    \end{subfigure}
    \caption{Comparison of performances on the body plans of learnable and fixed run champions with learnable and fixed controllers. 
    The first and last columns in each plot show the results of the main co-optimization experiment, and the second and third columns show the results when the learnable controller is optimized from scratch on body plans of learnable run champions and fixed run champions, respectively.
    In 3 out of 4 experimental settings, the body plans found by the fixed controller during the co-optimization experiments outperform the body plans found by the learnable controller when the learnable controller is optimized to control them. It demonstrates the failure of search over the morphology space during co-optimization with learnable controllers since they failed to discover high-performing body plans of fixed champions.
    }
    \label{fig:achieve_noobs}
\end{figure}
% \begin{figure}[th]
%     \centering
%     \begin{subfigure}{0.7\textwidth}
%         \includegraphics[width=\textwidth]{figures/boxplots/boxplot_bb55_walker_transfer_vs_decentralized.png}%
%     \end{subfigure}%
%     \begin{subfigure}{0.3\textwidth}
%         \includegraphics[width=0.8\textwidth]{figures/boxplots/boxplot_bb77_walker_transfer_vs_decentralized.png}\\
%         \includegraphics[width=0.8\textwidth]{figures/boxplots/boxplot_bb55_bridge_transfer_vs_decentralized.png}\\
%         \includegraphics[width=0.8\textwidth]{figures/boxplots/boxplot_bb77_bridge_transfer_vs_decentralized.png}%
%     \end{subfigure}
%     \caption{Distributions of the performances of different designs when the learnable controller is optimized to control them. The designs found by the fixed controller during the co-optimization experiments outperform the designs found by the learnable controller when the learnable controller is optimized to control them. It demonstrates the failure of search over the design space during co-optimization with learnable controllers since they failed to discover high-performing designs of fixed champions. 
%     % Each data point is plotted, and the mean values are marked with dark red and labeled. 
%     % Horizontal lines indicate statistically different results. ***: $P<0.001$, **: $P<0.005$, *: $P<0.05$
%     }
%     \label{fig:compare_designs}
% \end{figure}

Figure~\ref{fig:achieve_noobs} compares the performance of the learnable controller on the body plans of run champions found during co-optimization with learnable and morphology\-/only optimization with fixed controllers.  For comparison, we also show the original performances of the fixed run champions with the fixed controller. In 3 out of 4 experimental settings~(W5, W7, and B5; all $P < 0.05$), the learnable controller achieves significantly better performance on the body plans of fixed champions, which definitively demonstrates the existence of better body plans than the ones found during co-optimization with the learnable controller. \textit{To the best of our knowledge, this is the most concrete example to date showing the failure of brain-body co-optimization to find specific and demonstratedly-reachable high-performing regions of the morphology space.}%, demonstrating the poor search over the morphology space.}

It is also possible that simply the act of re-training the controller once the body plan has already been found and fixed produces a much different optimization trajectory than co-optimizing both at once.  To account for this we also take the run champions found when co-optimizing morphology with a learnable controller, freeze the resulting morphology and re-train a learnable controller on it from scratch (emulating the setup described above). Across all conditions, this resulted in statistically indistinguishable performance from the original co-optimized robots~(Figure~\ref{fig:achieve_noobs}, first versus second columns, all $P > 0.44$), demonstrating that the controllers found during co-optimization were well already converged to their current body plan.  This finding also provides further evidence consistent with the notion that the struggle to co-optimize brain-body systems is due to sub-par morphologies more so than sub-par controllers.  

For completeness, we also measured the performance of learnable run champions' body plans when they were controlled by the fixed controller. As expected, they performed very poorly (mean performance in W5: 4.88, W7: 6.31, B5: 2.64, B7: 2.47) and most solutions couldn't even reach the end of the environment. This suggests that the fixed controllers are not inherently better than the learnable controllers, but benefit from enabling search to discover more effective body plans, and especially body plans that are well adapted to the given control policy.  
% Additionally, we retrained learnable controllers from scratch for 5000 generations on the designs of learnable run champions, which resulted in similar performance distributions that are not statistically different~(all $P > 0.44$), demonstrating that the controllers were actually converged during the co-optimization process. 

Lastly, contrary to what we expect, we observe that the learnable controller, when it is trained on fixed run champions' body plans, doesn't always achieve the same level of performance as fixed controllers do (Figure~\ref{fig:achieve_noobs}, third versus last columns, $P < 0.05$ in W5, W7, and B7), which we interpret as the challenge of achieving very simple behaviors with complex neural network controllers when they are optimized through fitness alone. Future work could consider experimenting with more complex tasks and environments to understand how the co-optimization process unfolds when the complexity of the task/environment better matches the complexity of the controller.

\section{Discussion} 
\label{sect:discussion}

% Our co-optimization experiment and the following analysis produced a number of important results that can be seen in Figure~\ref{fig:achieve_noobs}. The first and last column in each plot shows the distribution of run champions from the co-optimization experiment with the learnable controller and the fixed controller, respectively, where for the latter we only optimize the design due to lack of optimizable parameters for the controller. The middle columns show the results of our analysis where we train learnable controllers on the designs of the fixed run champions. 

% To begin with, we have shown in our analysis that optimizing learnable controllers on the high-performing designs found with fixed controllers, resulted in better solutions compared to the solutions found with learnable controllers during co-optimization in most experimental settings (Figure~\ref{fig:achieve_noobs}, first vs middle columns, $P < 0.05$ in 3 out of 4). This result clearly demonstrates the failure of the search over the design space with learnable controllers during co-optimization, providing more evidence for the premature convergence of morphology as the main challenge of brain-body co-optimization. 

Our co-optimization experiment and the following analysis produced a number of important results that can be seen in Figure~\ref{fig:achieve_noobs}. Most importantly, we have shown that optimizing learnable controllers on the high-performing body plans found with fixed controllers, resulted in better solutions compared to the solutions found with learnable controllers during co-optimization in most experimental settings (first vs middle columns, $P < 0.05$ in 3 out of 4), which clearly demonstrates the failure of the search over the morphology space with learnable controllers during co-optimization, providing more evidence for the premature convergence of morphology as the main challenge of brain-body co-optimization.

While Cheney et al.~\cite{cheney_difficulty_2016} provides an embodied cognition perspective into this phenomenon, here we try to understand it from an optimization perspective. 
% Our experiments showing superior performance for body plan optimization with fixed controller compared to co-optimization of body plan and control, and the results of Mertan and Cheney~\cite{mertan_modular_2023} showing the weak response of learnable controllers to morphological changes provide an important insight into why the co-optimization process struggles to search the morphology space and challenge our conceptualization of fitness landscape. 
The standard intuition conceives a solution space where each solution has a single corresponding fitness creating a landscape over the solution space. However rugged it may be, we hope solutions with similar fitness values are grouped together in the solution space at some resolution, which allows us to search the space with heuristic search methods. This is certainly applicable to the brain-body co-optimization problem as well, where the brain and body together create the solution space. 
But instead of visualizing a standard landscape with one fitness value per brain-body pair, we could alternatively conceptualize two solution domains: one a fitness landscape over controllers and another a fitness landscape over body plans.  In doing so, we notice a peculiarity about brain-body solution space. While small changes in one part of the solution, that being the brain, can result in solutions with similar fitness and thus produce a typical and relatively smooth fitness landscape, changes to a robot's body almost always result in a severe fitness drop~\cite{mertan_modular_2023}. In light of this insight, we advocate for a new morphology-centric way of conceptualizing the fitness landscape that the co-optimization process runs on. \textit{Rather than imagining a fixed fitness value per each solution candidate, we can conceptualize the fitness landscape as a surface over the morphology space with a range of possible fitness values per body plan, where the actual fitness depends on the controller.} 

This new view allows us to understand the premature convergence of morphology from a search perspective. Body plans that we found early in the co-optimization process accrue more controller optimization steps, climbing up to their maximum fitness value -- an example of a first-mover advantage~\cite{lieberman1988first}. Conversely, body plans we found later on during the search via mutation, however promising these body plans may be, are at a disadvantage: they can't compete with first-movers until a controller is optimized for them to provide a better assessment of their maximum fitness, but typically can't survive long enough to accrue control optimization if they can't compete with first movers initially.

Indeed, Cheney et al.~\cite{cheney_difficulty_2016} investigate this by designing a smaller search space for the control which reduces the range of possible fitness values per body plan, effectively limiting the first-movers advantage, and showing that the premature convergence of morphology disappears. We take this even further by using a fixed controller which collapses the range of possible fitness values per body plan into a single value, turning the fitness landscape into a familiar format that we can search effectively. The success of the fixed controller shows that our algorithms are adequate in searching the standard landscape, but confronted with the peculiar co-optimization landscape that is unforgiving to changes in the body plan, they fail and get stuck in local optima. 

Luckily, this new understanding of the fitness landscape allows us to put recently proposed solutions into perspective: \textit{if the search over the morphology space is failing due to our inability to effectively measure the true potential of body plans, then (1) we could give our algorithms more resources to assess new body plans~\cite{cheney_scalable_2018}, and/or (2) we could try to have a better initial assessment for new body plans by alleviating the performance decline during the morphological search by using different genetic representations for coordinated changes~\cite{tanaka_co-evolving_2022,veenstra_effects_2022} or by using controllers that are robust to changes in the morphology~\cite{mertan_modular_2023}.} These two ways are orthogonal to each other and we believe further research on these avenues holds the potential for unlocking successful brain-body co-optimization.

It also enables us to draw connections to other fields.  For example, considering the optimization of the controller for a given morphology to be an approximation of the true potential (i.e. maximum) fitness of that morphology.  Thus our framework connects brain-body co-optimization with a wealth of literature on surrogate proxy fitness models~\cite{shi2010survey,jin2011surrogate}, on approaches to optimization under uncertainty such as for the multi-arm bandit problems~\cite{agrawal2012analysis, kuleshov2014algorithms}.  In robot brain-body co-optimization the computation expense of fitness evaluations makes it unreasonable to train individual controllers from scratch to ideally and independently determine the fitness of each morphology.  This is also the case for training the weights for a deep neural network architecture, and this framing draws analogies to the notion of sharing controllers for all possible body-plans/architectures in the fitness landscape, deemed supernet weight-sharing algorithms in the Neural Architecture Search literature~\cite{darts, survey2}.  It also suggests the importance of methods to enable effective transfer of controllers from one morphology to another when sharing these fitness approximators across disparate body plans via approaches like meta-learning, transfer learning, and few-shot learning~\cite{yosinski2014transferable, finn2017model, zhao2021few, mertan_modular_2023}.

Moreover, the success of body optimization with fixed controllers (Figure~\ref{fig:achieve_noobs}, first vs last columns, all $P < 0.005$) brings the formulation of ``brain" and ``body" into question. Formulating the control of the robot in ways that it assigns a single or smaller range of fitness values unlocks the body plan optimization by collapsing the fitness landscape and limiting the first-movers advantage. Solutions where robots' materials react to their environment in simpler ways and complex behavior emerges by exploiting the dynamics of the body and interaction with the environment, such as~\cite{creative_machines_lab_cornell_university_ithaca_ny_evolved_2014,cheney_unshackling_2014,cheney_evolving_2015,corucci_material_2016,corucci_evolving_2016}, can be further investigated. Especially, scaling this methodology to solve more complex tasks that would normally assumed to require complex closed-loop decision making should be investigated.

\section{Conclusion}

We have compared learnable neural network controllers and heuristically designed fixed controllers in the problem of co-optimization of brain and body (while for the latter, it is just the optimization of the body) for the locomotion task in multiple environments and morphology spaces that vary in complexity. Our experiments demonstrated the existence of high-performing regions that weren't found during the co-optimization process. Based on our analysis with the fixed controller, we developed a new morphology-centric understanding of the fitness landscape which explains the premature convergence from a search perspective and clarifies previously proposed solutions. We hope this new framework will help us systematically tackle the challenges of brain-body co-optimization.

\subsubsection{Acknowledgements.}

This material is based upon work supported by the National Science Foundation under Grant No. 2008413, 2239691, 2218063.
Computations were performed on the Vermont Advanced Computing Core supported in part by NSF Award No. OAC-1827314.

% \bibliography{all}
% \printbibliography
\bibliographystyle{splncs04}
\bibliography{all}

\begin{thebibliography}{10}
\providecommand{\url}[1]{\texttt{#1}}
\providecommand{\urlprefix}{URL }
\providecommand{\doi}[1]{https://doi.org/#1}

\bibitem{agrawal2012analysis}
Agrawal, S., Goyal, N.: Analysis of {T}hompson sampling for the multi-armed bandit problem. In: Conference on learning theory. pp. 39--1. JMLR Workshop and Conference Proceedings (2012)

\bibitem{bhatia2021evolution}
Bhatia, J., Jackson, H., Tian, Y., Xu, J., Matusik, W.: Evolution gym: A large-scale benchmark for evolving soft robots. Advances in Neural Information Processing Systems  \textbf{34},  2201--2214 (2021)

\bibitem{cheney_difficulty_2016}
Cheney, N., Bongard, J., Sunspiral, V., Lipson, H.: On the {Difficulty} of {Co}-{Optimizing} {Morphology} and {Control} in {Evolved} {Virtual} {Creatures}. In: Proceedings of the {Artificial} {Life} {Conference} 2016. pp. 226--233. MIT Press, Cancun, Mexico (2016). \doi{10.7551/978-0-262-33936-0-ch042}

\bibitem{cheney_evolving_2015}
Cheney, N., Bongard, J., Lipson, H.: Evolving {Soft} {Robots} in {Tight} {Spaces}. In: Proceedings of the 2015 {Annual} {Conference} on {Genetic} and {Evolutionary} {Computation}. pp. 935--942. ACM, Madrid Spain (Jul 2015). \doi{10.1145/2739480.2754662}

\bibitem{cheney_scalable_2018}
Cheney, N., Bongard, J., SunSpiral, V., Lipson, H.: Scalable co-optimization of morphology and control in embodied machines. Journal of The Royal Society Interface  \textbf{15}(143),  20170937 (Jun 2018). \doi{10.1098/rsif.2017.0937}

\bibitem{cheney_unshackling_2014}
Cheney, N., MacCurdy, R., Clune, J., Lipson, H.: Unshackling {Evolution}: {Evolving} {Soft} {Robots} with {Multiple} {Materials} and a {Powerful} {Generative} {Encoding} p.~8 (Aug 2014)

\bibitem{corucci_evolutionary_2017}
Corucci, F., Cheney, N., Kriegman, S., Bongard, J., Laschi, C.: Evolutionary {Developmental} {Soft} {Robotics} {As} a {Framework} to {Study} {Intelligence} and {Adaptive} {Behavior} in {Animals} and {Plants}. Frontiers in Robotics and AI  \textbf{4}, ~34 (2017). \doi{10.3389/frobt.2017.00034}

\bibitem{corucci_material_2016}
Corucci, F., Cheney, N., Lipson, H., Laschi, C., Bongard, J.: Material properties affect evolutions ability to exploit morphological computation in growing soft-bodied creatures. In: Proceedings of the {Artificial} {Life} {Conference} 2016. pp. 234--241. MIT Press, Cancun, Mexico (2016). \doi{10.7551/978-0-262-33936-0-ch043}

\bibitem{corucci_evolving_2016}
Corucci, F., Cheney, N., Lipson, H., Laschi, C., Bongard, J.C.: Evolving swimming soft-bodied creatures p.~2 (2016)

\bibitem{creative_machines_lab_cornell_university_ithaca_ny_evolved_2014}
{Creative Machines Lab, Cornell University. Ithaca, NY}, Cheney, N., Clune, J., Lipson, H.: Evolved {Electrophysiological} {Soft} {Robots}. In: Artificial {Life} 14: {Proceedings} of the {Fourteenth} {International} {Conference} on the {Synthesis} and {Simulation} of {Living} {Systems}. pp. 222--229. The MIT Press (Jul 2014). \doi{10.7551/978-0-262-32621-6-ch037}

\bibitem{ferigo_evolving_2021}
Ferigo, A., Iacca, G., Medvet, E., Pigozzi, F.: Evolving {Hebbian} {Learning} {Rules} in {Voxel}-based {Soft} {Robots}. preprint (Dec 2021). \doi{10.36227/techrxiv.17091218.v1}

\bibitem{finn2017model}
Finn, C., Abbeel, P., Levine, S.: Model-agnostic meta-learning for fast adaptation of deep networks. In: International conference on machine learning. pp. 1126--1135. PMLR (2017)

\bibitem{hiller_automatic_2012}
Hiller, J., Lipson, H.: Automatic {Design} and {Manufacture} of {Soft} {Robots}. IEEE Transactions on Robotics  \textbf{28}(2),  457--466 (Apr 2012). \doi{10.1109/TRO.2011.2172702}

\bibitem{hiller_dynamic_2014}
Hiller, J., Lipson, H.: Dynamic {Simulation} of {Soft} {Multimaterial} {3D}-{Printed} {Objects}. Soft Robotics  \textbf{1}(1),  88--101 (Mar 2014). \doi{10.1089/soro.2013.0010}

\bibitem{hiller_evolving_2010}
Hiller, J.D., Lipson, H.: Evolving amorphous robots. In: Alife. pp. 717--724. Citeseer (2010)

\bibitem{horibe_severe_2022}
Horibe, K., Walker, K., Berg~Palm, R., Sudhakaran, S., Risi, S.: Severe damage recovery in evolving soft robots through differentiable programming. Genetic Programming and Evolvable Machines  \textbf{23}(3),  405--426 (2022)

\bibitem{horibe_regenerating_2021}
Horibe, K., Walker, K., Risi, S.: Regenerating soft robots through neural cellular automata. In: Genetic Programming: 24th European Conference, EuroGP 2021, Held as Part of EvoStar 2021, Virtual Event, April 7--9, 2021, Proceedings 24. pp. 36--50. Springer (2021)

\bibitem{huang_one_2020}
Huang, W., Mordatch, I., Pathak, D.: One policy to control them all: Shared modular policies for agent-agnostic control. In: International Conference on Machine Learning. pp. 4455--4464. PMLR (2020)

\bibitem{jin2011surrogate}
Jin, Y.: Surrogate-assisted evolutionary computation: Recent advances and future challenges. Swarm and Evolutionary Computation  \textbf{1}(2),  61--70 (2011)

\bibitem{joachimczak_artificial_2016}
Joachimczak, M., Suzuki, R., Arita, T.: Artificial {Metamorphosis}: {Evolutionary} {Design} of {Transforming}, {Soft}-{Bodied} {Robots}. Artificial Life  \textbf{22}(3),  271--298 (Aug 2016). \doi{10.1162/ARTL\_a\_00207}

\bibitem{kim2013soft}
Kim, S., Laschi, C., Trimmer, B.: Soft robotics: a bioinspired evolution in robotics. Trends in biotechnology  \textbf{31}(5),  287--294 (2013)

\bibitem{kriegman_scalable_2020}
Kriegman, S., Blackiston, D., Levin, M., Bongard, J.: A scalable pipeline for designing reconfigurable organisms. Proceedings of the National Academy of Sciences  \textbf{117}(4),  1853--1859 (Jan 2020). \doi{10.1073/pnas.1910837117}, publisher: National Academy of Sciences Section: Physical Sciences

\bibitem{kriegman_kinematic_2021}
Kriegman, S., Blackiston, D., Levin, M., Bongard, J.: Kinematic self-replication in reconfigurable organisms. Proceedings of the National Academy of Sciences  \textbf{118}(49),  e2112672118 (Dec 2021). \doi{10.1073/pnas.2112672118}

\bibitem{kriegman_how_2018}
Kriegman, S., Cheney, N., Bongard, J.: How morphological development can guide evolution. Scientific Reports  \textbf{8}(1),  13934 (Dec 2018). \doi{10.1038/s41598-018-31868-7}

\bibitem{kriegman_interoceptive_2018}
Kriegman, S., Cheney, N., Corucci, F., Bongard, J.C.: Interoceptive robustness through environment-mediated morphological development. In: Proceedings of the {Genetic} and {Evolutionary} {Computation} {Conference}. pp. 109--116. ACM, Kyoto Japan (Jul 2018). \doi{10.1145/3205455.3205529}

\bibitem{kriegman_scale_2021}
Kriegman, S., Mohammadi~Nasab, A., Blackiston, D., Steele, H., Levin, M., Kramer-Bottiglio, R., Bongard, J.: Scale invariant robot behavior with fractals. In: Robotics: {Science} and {Systems} {XVII}. Robotics: Science and Systems Foundation (Jul 2021). \doi{10.15607/RSS.2021.XVII.059}

\bibitem{kriegman_scalable_2019}
Kriegman, S., Nasab, A.M., Shah, D., Steele, H., Branin, G., Levin, M., Bongard, J., Kramer-Bottiglio, R.: Scalable sim-to-real transfer of soft robot designs. In: 2020 3rd IEEE international conference on soft robotics (RoboSoft). pp. 359--366. IEEE (2020)

\bibitem{kriegman_automated_2019}
Kriegman, S., Walker, S., S.~Shah, D., Levin, M., Kramer-Bottiglio, R., Bongard, J.: Automated {Shapeshifting} for {Function} {Recovery} in {Damaged} {Robots}. In: Robotics: {Science} and {Systems} {XV}. Robotics: Science and Systems Foundation (Jun 2019). \doi{10.15607/RSS.2019.XV.028}

\bibitem{kuleshov2014algorithms}
Kuleshov, V., Precup, D.: Algorithms for multi-armed bandit problems. arXiv preprint arXiv:1402.6028  (2014)

\bibitem{lehman_evolving_2011}
Lehman, J., Stanley, K.O.: Evolving a diversity of virtual creatures through novelty search and local competition. In: Proceedings of the 13th annual conference on {Genetic} and evolutionary computation - {GECCO} '11. p.~211. ACM Press, Dublin, Ireland (2011). \doi{10.1145/2001576.2001606}

\bibitem{lieberman1988first}
Lieberman, M.B., Montgomery, D.B.: First-mover advantages. Strategic management journal  \textbf{9}(S1),  41--58 (1988)

\bibitem{darts}
Liu, H., Simonyan, K., Yang, Y.: {DARTS}: Differentiable architecture search. In: International Conference on Learning Representations (2019), \url{https://openreview.net/forum?id=S1eYHoC5FX}

\bibitem{liu2020voxcraft}
Liu, S., Matthews, D., Kriegman, S., Bongard, J.: Voxcraft-sim, a gpuaccelerated voxel-based physics engine (2020)

\bibitem{van2008visualizing}
Van~der Maaten, L., Hinton, G.: Visualizing data using t-sne. Journal of machine learning research  \textbf{9}(11) (2008)

\bibitem{marzougui_comparative_2022}
Marzougui, D., Biondina, M.: A {Comparative} {Analysis} on {Genome} {Pleiotropy} for {Evolved} {Soft} {Robots} p.~4 (2022)

\bibitem{medvet_evolution_2020}
Medvet, E., Bartoli, A., De~Lorenzo, A., Fidel, G.: Evolution of distributed neural controllers for voxel-based soft robots. In: Proceedings of the 2020 {Genetic} and {Evolutionary} {Computation} {Conference}. pp. 112--120. ACM, Cancún Mexico (Jun 2020). \doi{10.1145/3377930.3390173}

\bibitem{medvet_2d-vsr-sim_2020}
Medvet, E., Bartoli, A., De~Lorenzo, A., Seriani, S.: {2D}-{VSR}-{Sim}: {A} simulation tool for the optimization of 2-{D} voxel-based soft robots. SoftwareX  \textbf{12},  100573 (Jul 2020). \doi{10.1016/j.softx.2020.100573}

\bibitem{medvet_biodiversity_2021}
Medvet, E., Bartoli, A., Pigozzi, F., Rochelli, M.: Biodiversity in evolved voxel-based soft robots. In: Proceedings of the {Genetic} and {Evolutionary} {Computation} {Conference}. pp. 129--137. ACM, Lille France (Jun 2021). \doi{10.1145/3449639.3459315}

\bibitem{medvet_impact_2022}
Medvet, E., Nadizar, G., Pigozzi, F.: On the {Impact} of {Body} {Material} {Properties} on {Neuroevolutionfor} {Embodied} {Agents}: the {Case} of {Voxel}-based {Soft} {Robots} p.~9 (2022)

\bibitem{mertan_modular_2023}
Mertan, A., Cheney, N.: Modular {Controllers} {Facilitate} the {Co}-{Optimization} of {Morphology} and {Control} in {Soft} {Robots}. In: Proceedings of the {Genetic} and {Evolutionary} {Computation} {Conference}. pp. 174--183. ACM, Lisbon Portugal (Jul 2023). \doi{10.1145/3583131.3590416}

\bibitem{pfeifer2006body}
Pfeifer, R., Bongard, J.: How the body shapes the way we think: a new view of intelligence (2006)

\bibitem{pfeifer_morphological_2009}
Pfeifer, R., Iida, F.: Morphological computation: connecting body, brain, and environment p.~5 (2009)

\bibitem{pigozzi_evolving_2022}
Pigozzi, F., Tang, Y., Medvet, E., Ha, D.: Evolving modular soft robots without explicit inter-module communication using local self-attention. In: Proceedings of the Genetic and Evolutionary Computation Conference. pp. 148--157 (2022)

\bibitem{pontes-filho_single_2022}
Pontes-Filho, S., Walker, K., Najarro, E., Nichele, S., Risi, S.: A {Single} {Neural} {Cellular} {Automaton} for {Body}-{Brain} {Co}-evolution p.~4 (2022)

\bibitem{survey2}
Ren, P., Xiao, Y., Chang, X., Huang, P.y., Li, Z., Chen, X., Wang, X.: A comprehensive survey of neural architecture search: Challenges and solutions. ACM Comput. Surv.  \textbf{54}(4) (may 2021). \doi{10.1145/3447582}

\bibitem{rus2015design}
Rus, D., Tolley, M.T.: Design, fabrication and control of soft robots. Nature  \textbf{521}(7553),  467--475 (2015)

\bibitem{schmidt_age-fitness_2010}
Schmidt, M.D., Lipson, H.: Age-fitness pareto optimization p.~2 (2010)

\bibitem{shah_soft_2021}
Shah, D.S., Powers, J.P., Tilton, L.G., Kriegman, S., Bongard, J., Kramer-Bottiglio, R.: A soft robot that adapts to environments through shape change. Nature Machine Intelligence  \textbf{3}(1),  51--59 (2021)

\bibitem{shepherd2011multigait}
Shepherd, R.F., Ilievski, F., Choi, W., Morin, S.A., Stokes, A.A., Mazzeo, A.D., Chen, X., Wang, M., Whitesides, G.M.: Multigait soft robot. Proceedings of the national academy of sciences  \textbf{108}(51),  20400--20403 (2011)

\bibitem{shi2010survey}
Shi, L., Rasheed, K.: A survey of fitness approximation methods applied in evolutionary algorithms. In: Computational intelligence in expensive optimization problems, pp. 3--28. Springer (2010)

\bibitem{sims_evolving_1994}
Sims, K.: Evolving virtual creatures. In: Proceedings of the 21st annual conference on {Computer} graphics and interactive techniques - {SIGGRAPH} '94. pp. 15--22. ACM Press, Not Known (1994). \doi{10.1145/192161.192167}

\bibitem{talamini_evolutionary_2019}
Talamini, J., Medvet, E., Bartoli, A., Lorenzo, A.D.: Evolutionary {Synthesis} of {Sensing} {Controllers} for {Voxel}-based {Soft} {Robots} p.~8 (2019)

\bibitem{talamini_criticality-driven_2021}
Talamini, J., Medvet, E., Nichele, S.: Criticality-{Driven} {Evolution} of {Adaptable} {Morphologies} of {Voxel}-{Based} {Soft}-{Robots}. Frontiers in Robotics and AI  \textbf{8} (2021), \url{https://www.frontiersin.org/article/10.3389/frobt.2021.673156}

\bibitem{tanaka_co-evolving_2022}
Tanaka, F., Aranha, C.: Co-evolving morphology and control of soft robots using a single genome. In: 2022 IEEE Symposium Series on Computational Intelligence (SSCI). pp. 1235--1242. IEEE (2022)

\bibitem{trimmer_new_2008}
Trimmer, B.A.: New challenges in biorobotics: {Incorporating} soft tissue into control systems. Applied Bionics and Biomechanics  \textbf{5}(3),  119--126 (Dec 2008). \doi{10.1080/11762320802617255}

\bibitem{veenstra_how_2020}
Veenstra, F., Glette, K.: How {Different} {Encodings} {Affect} {Performance} and {Diversification} when {Evolving} the {Morphology} and {Control} of {2D} {Virtual} {Creatures}. In: The 2020 {Conference} on {Artificial} {Life}. pp. 592--601. MIT Press, Online (2020). \doi{10.1162/isal\_a\_00295}

\bibitem{veenstra_effects_2022}
Veenstra, F., Olsen, M.H., Glette, K.: Effects of {Encodings} and {Quality}-diversity on {Evolving} {2D} {Virtual} {Creatures} p.~4 (2022)

\bibitem{wilcoxon1964some}
Wilcoxon, F., Wilcox, R.A.: Some rapid approximate statistical procedures. (No Title)  (1964)

\bibitem{yim_modular_2007}
Yim, M., Shen, W.m., Salemi, B., Rus, D., Moll, M., Lipson, H., Klavins, E., Chirikjian, G.S.: Modular self-reconfigurable robot systems [grand challenges of robotics]. IEEE Robotics \& Automation Magazine  \textbf{14}(1),  43--52 (2007). \doi{10.1109/MRA.2007.339623}

\bibitem{yosinski2014transferable}
Yosinski, J., Clune, J., Bengio, Y., Lipson, H.: How transferable are features in deep neural networks? Advances in neural information processing systems  \textbf{27} (2014)

\bibitem{zhao2021few}
Zhao, Y., Wang, L., Tian, Y., Fonseca, R., Guo, T.: Few-shot neural architecture search. In: International Conference on Machine Learning. pp. 12707--12718. PMLR (2021)

\end{thebibliography}

\end{document}